# Enhancing Face Recognition with Latent Space Data Augmentation and Facial Posture Reconstruction

Soroush Hashemifar[1], Abdolreza Marefat[2], Javad Hassannataj Joloudari[3,*], Hamid Hassanpour[4]

[1]School of Computer Engineering, Iran University of Science and Technology, Tehran, Iran. hashemifar_soroush@cmps2.iust.ac.ir
[2]Department of Artificial Intelligence, Technical and Engineering Faculty, South Tehran Branch, Islamic Azad University, Tehran, Iran. rzamarefat@gmail.com
[3]Department of Computer Engineering, Faculty of Engineering, University of Birjand, Birjand 9717434765, Iran
[4]Faculty of Computer Engineering & Information Technology, Shahrood University of Technology, P.O. Box 316, Shahrood, Iran. h.hassanpour@shahroodut.ac.ir

*Correspondence: javad.hassannataj@birjand.ac.ir

**Abstract**

The small amount of training data for many state-of-the-art deep learning-based Face Recognition (FR) systems causes a marked deterioration in their performance. Although a considerable amount of research has addressed this issue by inventing new data augmentation techniques, using either input space transformations or Generative Adversarial Networks (GAN) for feature space augmentations, these techniques have yet to satisfy expectations. In this paper, we propose an approach named the Face Representation Augmentation (FRA) for augmenting face datasets. To the best of our knowledge, FRA is the first method that shifts its focus towards manipulating the face embeddings generated by any face representation learning algorithm to create new embeddings representing the same identity and facial emotion but with an altered posture. Extensive experiments conducted in this study convince of the efficacy of our methodology and its power to provide noiseless, completely new facial representations to improve the training procedure of any FR algorithm. Therefore, FRA can help the recent state-of-the-art FR methods by providing more data for training FR systems. The proposed method, using experiments conducted on the Karolinska Directed Emotional Faces (KDEF) dataset, improves the identity classification accuracies by 9.52 %, 10.04 %, and 16.60 %, in comparison with the base models of MagFace, ArcFace, and CosFace, respectively.

# 1. Introduction

Face images are widely used as a biometric modality in Face Recognition (FR) systems, making them highly popular (Terhorst, Kolf, Damer, Kirchbuchner, & Kuijper, 2020). It is used in a wide range of contexts with the aim of identity authentication, and its applications vary from daily life and finance to military and public security (M. Wang & Deng, 2021). In fact, in comparison with other biometrics, such as the fingerprint, iris, or retina, which are ubiquitously used for authorizing individuals, FR can provide us with the most convenient way to capture visual information without needing any extra activity from the subject. In recent years, FR has been one of the most proactively studied areas in Computer Vision (Ali, Tian, Din, Iradukunda, & Khan, 2021). Mainly, with the advent of deep learning and architectures like Convolutional Neural Networks (CNNs) (Krizhevsky, Sutskever, & Hinton, 2017), a large number of efficient facial recognition methods with outstanding performance have been invented to address this challenge (Bhattacharyya, Chatterjee, Sen, Sinitca, Kaplun, & Sarkar, 2021; Cao, Shen, Xie, Parkhi, & Zisserman, 2018; Meng, Zhao, Huang, & Zhou, 2021; Peng, Wang, Li, & Gao, 2019; Taigman, Yang, Ranzato, & Wolf, 2014; Yan, et al., 2022; Zhang, Fang, Wen, Li, & Qiao, 2017; Y. Zheng, Pal, & Savvides, 2018). These successful algorithms depend heavily on the performance of neural networks, which use a cascade of layers comprised of neurons that can learn different levels of abstractions and representations from the input data (M. Wang & Deng, 2021). These representations are more powerful substitutions for hand-crafted features from facial attributes such as Scale-Invariant Feature Transform (SIFT) and Speeded Up Robust Features (SURF) (Gupta, Thakur, & Kumar, 2021; Kortli, Jridi, Al Falou, & Atri, 2018). Their main benefit is that they eliminate the necessity of manually and thoroughly searching for the best features representing one's face. Moreover, learning representations via deep learning-based algorithms makes the generated features surprisingly discriminative in that the network considers inter-class diversity and intra-class compactness within the training data (Duan, Lu, & Zhou, 2019).

However, there are still problematic scenarios in which FR systems fail to realize the expectations. For instance, in real-life situations, the imagery of a person's face has a high chance of being in a variety of facial expressions, occlusions, poor illumination, low resolution, etc. (Lahasan, Lutfi, & San-Segundo, 2019; P. Li, Prieto, Mery, & Flynn, 2019; Y. Li, Guo, Lu, & Liu, 2021), and all these factors cause substantial degradation of the overall performance of the current FR algorithms. Thus, different approaches have been adopted to rectify the negative impact of such barriers in FR systems (Deng & Zafeririou, 2019b; Lu, Jiang, & Kot, 2018; Song, Gong, Li, Liu, & Liu, 2019). Some have opted for experimenting and devising new loss functions whose capability to better feedback to their neural network in the backpropagation step enables the extracted deep features to be more discriminative and separable (Deng, Guo, Xue, & Zafeiriou, 2019a; Y. Huang et al., 2020; Meng, Zhao et al., 2021; Srivastava, Murali, & Dubey, 2020; H. Wang et al., 2018; M. Wang & Deng, 2021; X. Wang, Wang, Wang, Shi, & Mei, 2019; Y. Zheng, Pal, & Savvides, 2018). In addition to these works, different architectures have been implemented to extract feature maps that are more useful for facial representations.

Moreover, developing larger and more variant datasets has been one of the primary stimuli pushing the boundaries in recent FR systems (Z. Liu, Luo, Wang, & Tang, 2015). Nevertheless, although some of these benchmark datasets can be found in large volumes, we often lack such a training set of images when for real use cases. A typical case would be a situation in which the goal is to train a deep learning-based method on a private, in-house set of identities chosen by a multimedia organization for video indexing purposes. The data-gathering phase can be very time-consuming and labor-consuming and sometimes even impossible, and it acts as an impediment in the way of achieving a tailored amount of training datasets. These have motivated researchers to pave the way by introducing different data augmentation techniques.

Data augmentation refers to a set of techniques to increase the number of training datasets without losing previously annotated data. The benefit of such methods is that they equip the trained model with more generalizability and act as a regularizer in the case of overfitting, which is one of the most frequent complications when dealing with a small amount of training data (Lv, Shao, Huang, Zhou, & Zhou, 2017; Wong, Gatt, Stamatescu, & McDonnell, 2016). Overall, there are two mainstream categories of methods for augmenting data. The first set of methods has the aim of manipulating the data in the input space in that they simply take the input image and apply different geometric transformations such as translations, cropping, vertical and horizontal flipping, rotation, etc. (Masi, Trần, Hassner, Sahin, & Medioni, 2019). Even though these methods are proven to be extremely useful in some other challenges like image classification, object detection, and image captioning in computer vision, in the case of FR, they cannot be as helpful as they expected. The main reason is that for any FR system to capture a reliable visual representation of a face crop image, the content should be aligned in terms of facial landmarks. This means that any geometric alteration on these, which conspicuously happens when one uses these classical methods, can perturb the overall performance of the FR pipeline. These challenges have motivated the researchers to shift their studies' direction toward more modern and domain-specific solutions (D. Jiang, Hu, Yan, Zhang, Zhang, & Gao, 2005; Mohammadzade & Hatzinakos, 2012; Shen, Luo, Yan, Wang, & Tang, 2018), leading to the second set of methods, which are known to be Generative Adversarial Networks (GANs) (Goodfellow et al., 2020). These methods are the well-known type of generative models that are used to transform the input data in feature space to generate new augmented image data. This collection of models has the ability to modify the facial characteristics present in an image of a face, such as the hairstyle, facial expression, body posture, skin color, and so on, in order to match a desired style. However, in most cases, these generative models cannot create realistic outputs, and these models deal with the high complexities of reconstructing the feature space to input space without having any considerable improvement on the downstream task, which in our case, is classification on the identity of the samples.

In this paper, we present the Face Representation Augmentation (FRA) approach as a solution to these challenges. This approach augments the posture of a given face image in the latent space. This means that, given a set of embeddings representing a specific person, the proposed approach alters the embedding to sustain the identity-related features with a transformed pose feature. The FRA approach can help the existing facial recognition systems, especially when the number of training samples is imbalanced or less than expected. Our main contributions in this paper are itemized in the following:

1. An approach is proposed to augment facial posture within the latent space, aiming to simplify the complexity of the image augmentation issue.
2. Generating embeddings that are noise-free and non-duplicate, and have been demonstrated to be linearly separable.
3. Extensive experiments were conducted on the Karolinska Directed Emotional Faces (KDEF) (Lundqvist, Flykt, & Öhman, 1998) dataset and they improved the identity classification accuracies in comparison with the base models of MagFace, ArcFace, and CosFace, respectively.

The rest of the paper is organized as follows. In Section 2, we will provide a brief overview of the relevant studies conducted on face-specific data augmentation and representation learning. Then, in Section 4, we present the details of our proposed methodology. In Section 4, we demonstrate the results of our experiments in comparison with other related state-of-the-art approaches. Finally, the conclusion will be drawn in Section 5.

## 2. Related works

In this section, we present an overview of face-specific data augmentation techniques. These are categorized into two groups: classical and generative-based methods in 2.1. Additionally, we review the related literature of FR algorithms in 2.2.

### 2.1. Face-Specific Data Augmentation

To begin, five data augmentation techniques for face photos were reported by Lv et al. (2017). These techniques were landmark perturbation, hairdo synthesis, glasses synthesis, postures synthesis, and lighting synthesis. Vincent et al. (2010) tried to synthesize more data by applying different types of noise, such as Gaussian and Salt-and-pepper with the objective of training Stacked Denoising Autoencoders on more complicated samples. Wang et al. ( 2017) addressed the issue of data augmentation in picture classification using conventional transformation techniques and GANs. They also suggested a technique for learning network-based augmentations that better enhance the classifier in the setting of generic photos rather than face images.

Moreover, although hair is not an intrinsic part of the human face, it interferes with facial recognition since it obscures the face and changes its appearance. The DiscoGAN architecture, devised by Kim et al. (2017), aims to identify connections between domains by utilizing unpaired data. Its primary objective is to facilitate the modification of hair color. In addition to the color, they suggested changing the bang by transferring an unsupervised visual characteristic using a reconfigurable GAN. An online compositing technique was used in the face synthesis system proposed by Kemelmacher-Shlizerman et al. (2016). The system might produce a series of fresh photographs with the input person's identification and the questioned look using one or more photos of their face and a text query like curly hair. In another study, the authors have proposed Pose and expression resilient Spatial aware GAN (PSGAN) (W. Jiang et al., 2020). It starts using the Makeup Distill Network to separate the reference image's makeup into two spatially aware makeup matrices. Following this, a module known as attentive makeup morphing has been developed to enable users to articulate the modifications made to a pixel's visual characteristics in the source image, influenced by the reference image. To ease applications in the real-world setting, PSGAN is the first to concurrently accomplish partial, shade tunable, and pose/expression robust makeup transfer. In order to separate the makeup from the reference picture as two makeup matrices, an MDNet is also included. The transfer that allows for flexibility, partial shading adjustment, and spatial awareness is made possible by the makeup matrices. To acquire knowledge about all the characteristics of cosmetics (Nguyen, Tran, & Hoai, 2021), such as color, form, texture, and position, it incorporates an improved branch for color transfer and a novel branch for pattern transfer. They present makeup in this work as a combination of color transformation and pattern addition, and they create a thorough makeup transfer technique that works for both delicate and dramatic looks. They suggest using warped faces in the Ultraviolet (UV) space while training two network branches to eliminate the disagreement between input faces regarding form, head posture, and expression. They also developed a new architecture consisting of two branches dedicated to the transfer of color and pattern. They present brand-new cosmetics transfer datasets with extreme fashions that were not considered in the earlier datasets.

## 2.2. Representation Learning for Face Recognition

Representation learning refers to a set of algorithms that are designed to solve a variety of challenges like image retrieval (Movshovitz-Attias, Toshev, Leung, Ioffe, & Singh, 2017; Ustinova & Lempitsky, 2016; Wu, Manmatha, Smola, & Krahenbuhl, 2017), the person (Chen, Chen, Zhang, & Huang, 2017; Xiao, Luo, & Zhang, 2017) and vehicle (Bai, Lou, Gao, Wang, Wu, & Duan, 2018; Sanakoyeu, Tschernezki, Buchler, & Ommer, 2019) re-identification, landmark detection, and fine-grained object recognition (Em, Gag, Lou, Wang, Huang, & Duan, 2017; Smirnov, et al., 2019). The process of facial recognition in computer vision relies heavily on acquiring knowledge of patterns that possess distinct similarities within a class and significant differences between classes (Shi, Yu, Sohn, Chandraker, & Jain, 2020). Previous works (Deng et al., 2019; W. Liu, Wen, Yu, Li, Raj, & Song, 2017; Meng, Zhao et al., 2021; Schroff, Kalenichenko, & Philbin, 2015; H. Wang et al., 2018) have mainly adopted different, more robust loss functions to learn representations that satisfy the requirements above.

In (Schroff, Kalenichenko, & Philbin, 2015), a deep convolutional neural network named FaceNet, was proposed, which learns facial representations with the help of triplet loss. The main objective of this work is to achieve an embedding $f(x)$ from an image $x$ into a d-dimensional Euclidean space $R^d$. The obtained embedding is generated so that the squared distance among the embeddings from one class is small and that of the embeddings from different classes is large. This algorithm achieves 99.63% and 95.12% accuracy in LFW (G. B. Huang, Mattar, Berg, & Learned-Miller, 2008) and YouTube Faces Database (Wolf, Hassner, & Maoz, 2011), respectively. Liu et al. (2017) have proposed a new look at the loss functions based on the Euclidean margin between the produced embeddings. For CNNs to learn discriminative facial characteristics with clear and innovative geometric interpretation, they suggest the A-Softmax loss. The assumption that faces also lie on a manifold is fundamentally compatible with the learned features' discriminative spread on a hypersphere manifold. In order to approximate the learning problem that minimal inter-class distance is greater than maximum intra-class distance, they develop a lower margin set between such classes.

In (Deng et al., 2019), the authors have proposed ArcFace, a major modification of the Softmax loss to further improve the robustness of the learned deep features. By using the arc-cosine function to determine the angle between the current feture and the desired weight, and incorporating an additional angular margin to the desired angle, they obtained the desired logit. Then, these logits are rescaled by a fixed feature norm followed by the same steps in the Softmax loss function. Their approach has the following advantages over the others. (1) Directly optimizing the geodesic distance margin (2) State-of-the-art performance in several benchmark datasets: achieving 99.53% accuracy (3) Easiness in terms of implementation (4) Efficiency in terms of computational complexity.

In (H. Wang et al., 2018), the authors reformulated the Softmax loss as a cosine loss to introduce a novel loss function named Large Margin Cosine Loss (LMCL). Their improvement is to further maximize the decision margin in the angular space by introducing and training a deep model called CosFace. In this deep model, LMCL guides the convolutional layers to learn features with huge cosine margins. Their results demonstrate that they have achieved 97.96% accuracy in face verification on the MegaFace benchmark, which has been a significant improvement in comparison to previous works.

Meng et al. (2021) proposed a new set of losses that enable the network to learn embeddings whose magnitude represents the quality of the given face. By extending ArcFace (Deng et al., 2019) and introducing the MagFace loss function, they demonstrate that the more likely the subject is to be recognized, the bigger the magnitude of the generated embedding. MagFace acquires the ability to produce these universal embeddings by drawing the simpler examples from a group of identities towards the center of the group and moving them away from the origin. This makes the embeddings robust to ambiguity and the absence of high discriminative features which prevalently exist in unconstrained face images in real scenarios. They have achieved 99.83% verification accuracy in the LFW benchmark dataset. In Table 1, a comparison of these works is depicted.

**Table 1.** Verification accuracy of MagFace, CosFace, ArcFace, and SphereFace. These models are evaluated on CALFW, CPLFW, AgeDB, LFW, and CFP-FP datasets.

| Method | CALFW (T. Zheng, Deng, & Hu, 2017) | CPLFW (T. Zheng & Deng, 2018) | AgeDB (Moschoglou, Papaioannou, Sagonas, Deng, Kotsia, & Zafeiriou, 2017) | LFW (G. B. Huang et al., 2008) | CFP-FP (Sengupta, Chen, Castillo, Patel, Chellappa, & Jacobs, 2016) |
|---|---|---|---|---|---|
| MagFace (Meng, Zhao et al., 2021) | 96.15 | 92.87 | 98.17 | 99.83 | 98.46 |
| CosFace (H. Wang et al., 2018) | 96.18 | 92.18 | 98.17 | 99.78 | 98.26 |
| ArcFace (Deng et al., 2019) | 95.96 | 92.72 | 98.05 | 99.81 | 98.40 |
| SphereFace (W. Liu et al., 2017) | 95.58 | 91.27 | 97.05 | 99.67 | 96.84 |

Moreover, although these approaches have significant performance, directly applying GAN approaches appears to have a few disadvantages. Models collapse, difficulty in training and convergence problems, and poor image generation effect, along with the unreliable results of the generator for unconstrained input images (Tran, Yin, & Liu, 2017), cause the generated image examples to be incapable of being utilized for industrial data augmentation tasks (Cong & Zhou, 2022; Kammoun, Slama, Tabia, Ouni, & Abid, 2022). In addition, the majority of GAN networks require a significant amount of time to reach a state of convergence and produce satisfactory outcomes. This timeframe can vary, spanning from several thousand epochs to a duration of days or even months, as demonstrated by the methodology introduced in (Tran, Yin, & Liu, 2017). It is important to note that in GAN methods when we transform the face into smaller embeddings, certain details such as facial textures and shapes may be lost.

Therefore, reconstruction-based methods often have difficulty recovering these specific features (Tian, Peng, Zhao, Zhang, & Metaxas, 2018).

# 3. Proposed approach

This section presents the proposed FRA approach. As can be inferred from Figure 1, our method includes four steps. These are as follows: face detection and alignment, input preparation, pose feature extraction, and representation augmentation. Steps 1 and 2 comprise our data preprocessing pipeline explained in Section 3.1. Steps 3 and 4 represent our main contribution to this paper and are explained in Sections 3.2-3.4. Finally, Section 3.5 describes the loss function.

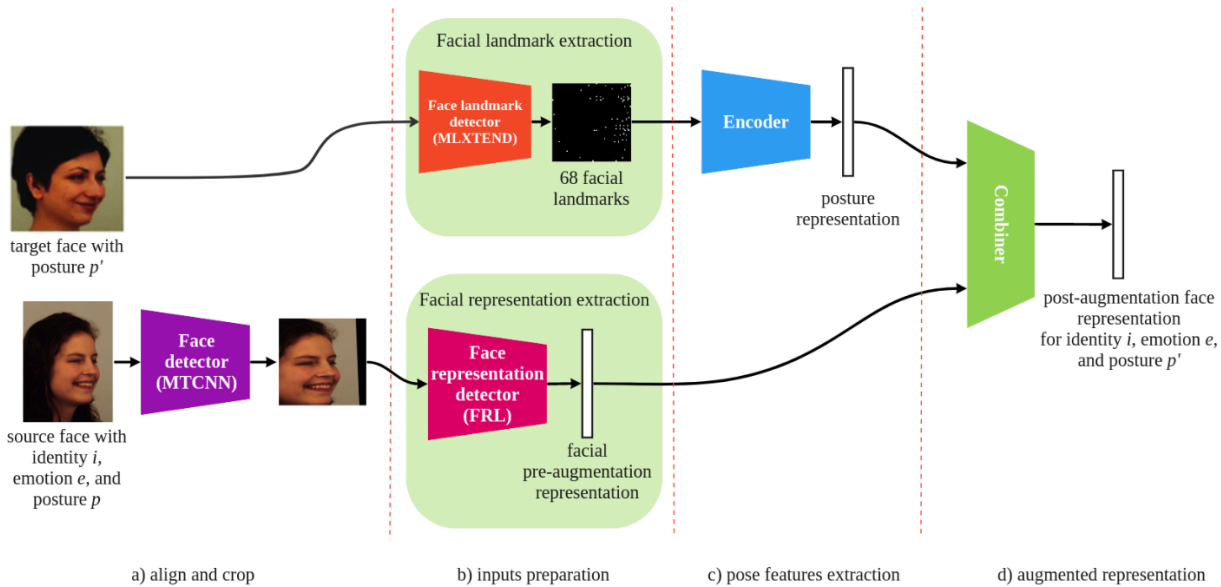

**Figure 1.** The overall procedure of FRA. FRA consists of four stages aimed at producing a new representation vector with identity *i*, emotion *e*, and posture *p*. These stages involve the application of a target posture *p* onto a base image with identity *i* and emotion *e*.

## 3.1. Dataset preprocessing and preparation

Our data preprocessing step includes three main phases. These three phases are depicted in Figure 1. As is seen in the first phase, we feed the raw face images to the Multi-task Cascaded Convolutional Networks (MTCNN) algorithm (W. Liu et al., 2017), which is a robust face and landmark detector. MTCNN provides five landmark points, including the center of both eyes, the tip of the nose, and the left and right corners of the lips, and a bounding box that perfectly encloses the face area within the image without any padding. In this phase, we also align the face images by feeding both the obtained facial landmarks and the face image itself to the warp affine method. This method is available in OpenCV (Bradski, 2000), a renowned library containing pre-existing algorithms related to computer vision.

In the second phase, we feed the aligned face images to MLXTEND[1] to determine more facial landmarks. As is shown in Figure 1, MLXTEND outputs 68 facial key points, which we use to construct binarized images with pixel value 0 (completely black) for the background and 1 (completely white) for facial landmarks. On the other hand, we need to have fixed-size embeddings for each sample within the dataset. The training data for the combiner module, which will be discussed in Section 3.4, consists of these embeddings. In our case, we use three of the most reliable and robust face representation learning (FRL) algorithms, namely MagFace (Meng Zhao et al., 2021) ArcFace (Deng et al., 2019a), and CosFace (H. Wang et al., 2018), for obtaining embedding of an input image. In Phase 3, the binarized images generated in Phase 2 are fed to the AE model in order to develop an embedding vector representing posture features. Finally, in Phase 4, pose and face representation vectors are fed into the combiner module to develop an augmented face representation vector.

## 3.2. Facial Landmark Restoration using Autoencoders

Autoencoders (AE) are a particular type of neural network whose primary functionality is to encode the input into a meaningfully compacted representation and decode this into the input space afterward (Bank, Koenigstein, & Giryes, 2020; Rumelhart, Hinton, & Williams, 1985). On the one hand, because the proposed method is designed to augment a face representation in the latent space, the augmentation is also generated as a representation vector. On the contrary, since FRL strives to augment the position of the face, we must convert the facial position into a vector that represents the position. In order to reach this posture representation, we have been inspired by the work done by Meng et al. (Meng, Xu et al., 2021) and decided to use an AE-based model for encoding our input space (binarized images of landmarks explained in Section 3.2) into the latent space (embeddings), as shown in Figure 2.

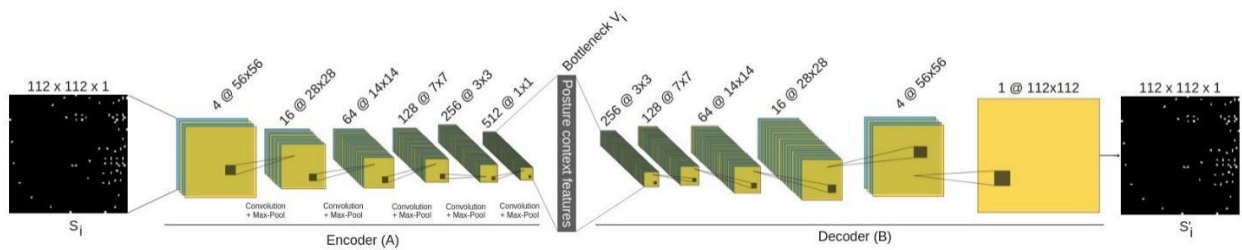

**Figure 2.** The general architecture of an autoencoder-based model. FRA utilizes a typical convolutional autoencoder. The output of the encoder, known as the bottleneck, is utilized in subsequent stages.

Given $S_i$ as a sample of facial landmarks image, the output of $F(S_i)$ is a reconstructed image $S'_i$, where indicates the autoencoder architecture. After the AE model's convergence, we can discard $B$ *(decoder)* and take only *A*, which has learned to encode the input landmarks' image into an optimized and meaningful latent space representation denoted by $V_i$. It is worth mentioning that $V_i$ plays a vital role in our proposed method, which is the latent representation of the posture of the face. $V_i$ represents the desired augmentation of posture that will transform the pre-augmented face representation in subsequent steps. Figure 2 illustrates the proposed AE-based model and its architecture.

---
[1] http://rasbt.github.io/mlxtend/

## 3.3. Combining Feature Vectors and Feature Extraction using Vision Transformers

Vision Transformers (ViT) are deep learning models whose versatility in various fields, such as natural language processing, speech recognition, and computer vision, has made them a prominent choice for researchers (Lin, Wang, Liu, & Qiu, 2022). In comparison with the conventional CNNs, ViT models have achieved competitively superior results in vision tasks like object detection (Carion, Massa, Synnaeve, Usunier, Kirillov, & Zagoruyko, 2020), image recognition (Touvron, Cord, Douze, Massa, Sablayrolles, & Jégou, 2021), image super-resolution (Yang, Yang, Fu, Lu, & Guo, 2020), and segmentation (Hafiz, Parah, & Bhat, 2021; Ye, Rochan, Liu, & Wang, 2019).

At the core of ViT models, there is a mechanism of attention that has been probably one of the most significant concepts deep learning. Its inspiration is the biological attributes of human beings in that, to recognize an object, we tend to focus on the most distinctive parts of that entity instead of paying attention to all parts of it as a whole (Niu, Zhong, & Yu, 2021). In terms of deep neural networks, this can be interpreted as assigning importance scores for a given set of features where the higher scores are for more relevant features and the lower ones for the features with less saliency (Chollet, 2021). As can be observed in Figure 3, the model demonstrates an increased emphasis on the parts depicting the target object in the image.

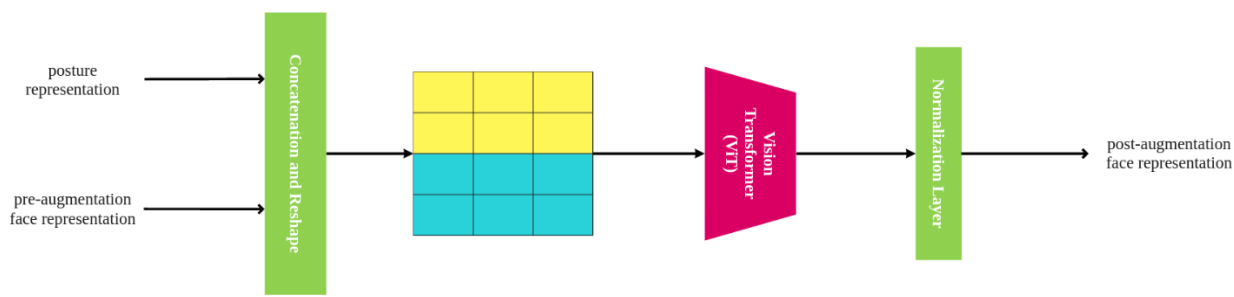

**Figure 3.** The paradigm of combining two representation vectors using ViT. The combiner takes two representation vectors and reshapes them into a matrix to be processed by a vision-transformer component.

Moreover, transformers (Vaswani et al., 2017) are a collection of neural networks that employ the attention mechanism. These models consist of multiple encoders and decoders whose architectures are identical. In these models, the input is encoded using a multi-head self-attention (MSA) mechanism, and then decoders are used to process the encoder's output, which includes an additional attention layer. Self-attention is a function denoted in Equation (1).

$$Attention(Q,K,V) = softmax\left(\frac{Q.K^T}{\sqrt{d_k}}\right).V \quad (1)$$
$$\text{s.t. } Q = W^Q x, K = W^K x, V = W^V x$$

Based on (1), $W^Q$, $W^K$, and $W^V$ are weight matrices used in linear transformations on inputs *x* to produce *Q*, *K*, and *V*. The attention score is then calculated by $Q.K^T$ as the dot product of the query and each key, scaled by the dimension $d_k$ of the key *K*. In MSA, *Q*, *K*, and *V* are projected linearly, and this is done for *h* consecutive times with different learned weights. Then, by applying the self-attention mechanism on each of the outputs in the previous step simultaneously, we obtain *h* outputs which are heads. Then, these heads are concatenated to achieve the final output. The following demonstrates these computations in mathematical terms.

$$MultiHeadAttention(Q,K,V) = Concat(head_1, head_2, \dots, head_h).W^O \quad (2)$$
$$\text{s.t. } head_i = Attention(QW^Q, KW^K, VW^V)$$

MSAs, in contrast to CNNs, utilize large data-specific kernels to transform feature maps, resulting in an equal level of expressiveness as the architectures based on CNNs (Cordonnier, Loukas, & Jaggi, 2019). The key difference exists where convolutions diversify feature maps, whereas MSAs combine them. According to (Park & Kim 2022b), the Fourier analysis of feature maps demonstrates that convolutions boost high-frequency components, whereas MSAs, on the other hand, attenuate them.

Furthermore, finding elements that are more pertinent for depicting the altered posture is made easier by the multi-head attention layer. In order to do this, the scaled dot product attention gives greater weight to the characteristics of the input facial representation and encoded posture that is more pertinent while providing less weight to the less relevant features (Park & Kim 2022a). The procedure chooses features from various input regions and aids in improving representation performance since there are several heads in the attention layer.

In this paper, we have used a ViT-based architecture for extracting features. As mentioned earlier, this policy guarantees that the model is trained to focus on the most significant feature values within the identity and posture-related feature vectors at the same time. To achieve this objective, two vectors that represent the encoder and the FRL are concatenated into one vector and then transformed into a matrix by reshaping. This matrix is then fed to the ViT module, which breaks down this representation matrix into patches, as discussed earlier, and aggregates the information extracted from each patch. To our knowledge, a representation vector consists of various numerical features of the input image. So, attending to each patch forces ViT to extract helpful information from more essential parts of these representation vectors, instead of considering whole elements of representations unthinkingly, which leads to a better-augmented representation.

## 3.4. Generating Pose-Aware Face Embeddings

In the concluding stage of our approach, the output is directed to a fully connected layer, followed by normalization. This results in a vector with an arbitrary size representing the augmented representation specific to the target face. We draw inspiration from the network architecture of ResNet100, which serves as the backbone for MagFace, ArcFace, and CosFace. The final normalization step is applied to learn the necessary statistics, ensuring that the generated representation aligns with the range of face representations and facilitates smoother convergence of the entire architecture. Consequently, the generated representation becomes more akin to the target representation.

## 3.5. Multi-Task Loss Function

During the training process, we employed a Multi-part Loss Function (MLF) as the learning objective. This MLF consists of a Binary Cross-Entropy (BCE) loss function, which is employed for training the autoencoder in order to enhance the reconstruction of the posed style. Since the activation function of the last layer of our autoencoder is *Sigmoid*, it can lead to loss of saturation (plateau) (Goodfellow, Bengio, & Courville, 2016). This saturation could prevent gradient-based learning algorithms from convergence. To prevent this problem, it would be more advantageous to incorporate a *logarithm function* into the objective function to reverse the effects of the *exponential function* present in the *Sigmoid*. *BCE* is the preferred choice because it utilizes a *logarithm function*, unlike Mean Squared Error (MSE).

The second part of our loss function is a type of *N-pair Loss* (Sohn, 2016). *N-pair loss* generalizes triplet loss (Schroff, Kalenichenko, & Philbin, 2015) to include comparison with multiple negative samples. The purpose of this function is to keep the distance between the anchor and positive less than the distance between the anchor and negative representations, as shown in Figure 4.

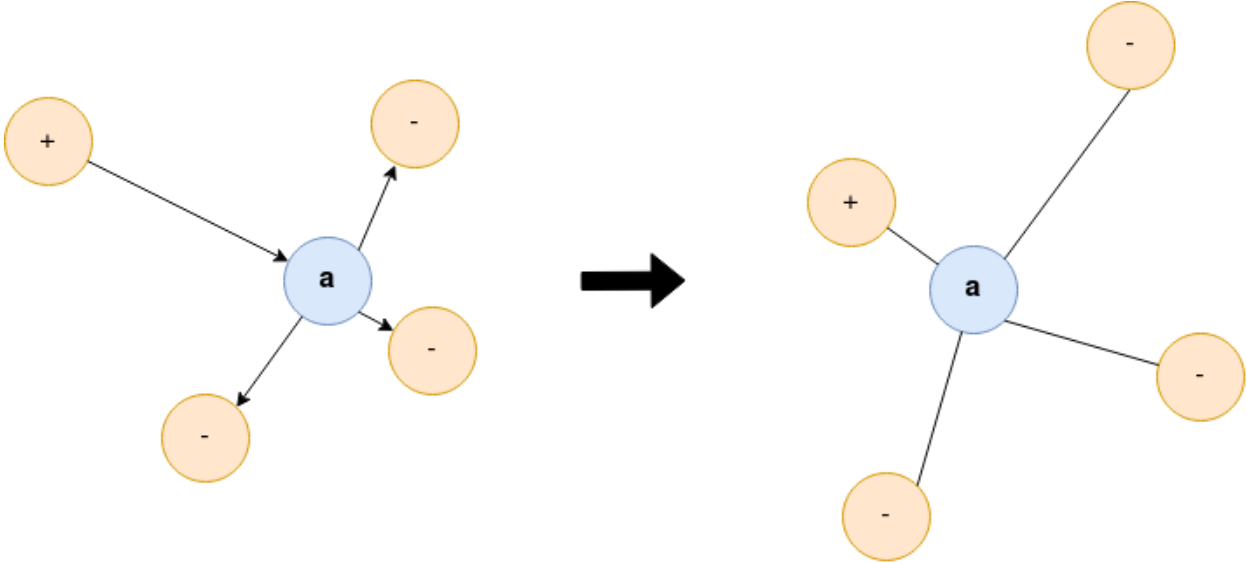

**Figure 4.** Impact of the proposed loss function on the learning process. The N-pair loss enables the model to discriminate as far as possible between representation vectors of pose variants with the same identity and emotion.

The proposed multi-task loss function is defined as follows:

$$L_{total} = -\sum_{i}^{N}(y_i \log\log(p_i) + (1-y_i)\log\log(1-p_i)) + L_t(X^a, X^p, X^n_{pose}) \quad (3)$$
$$+ L_t(X^a, X^p, X^n_{identity}) + L_t(X^a, X^p, X^n_{emotion})$$

According to (3), y$_i$ and p$_i$ denote the reconstructed pose style and the original pose style, respectively, and $L_t(X^a, X^p, X^n)$ is defined as follows:

$$L_t = \frac{1}{N} \sum_i^N max(d(f(X_i^a), f(X_i^p)) - d(f(X_i^a), f(X_i^n)) + m, 0) \qquad (4)$$

The margin, denoted by *m,* is used to establish a distinction between the distance of the anchor and positive pair, as well as the anchor and negative representation vectors. The function *f* represents the suggested architecture. *d* is the Euclidean distance applied on normalized features, and is given by Equation (5).

$$d(x_i, y_i) = \|x_i - y_i\|_2^2 \qquad (5)$$

In Equation (6), *a* and *p* denote the anchor (generated) representation and the positive (real) representation, respectively. In ddition, $X_{pose}^n$, $X_{identity}^n$, and $X_{emotion}^n$ represent the negative representation with respect to pose, identity, and emotion of the anchor face, respectively. Specifically, negative pose representations have the same identity as the anchor but with different poses. The same holds for negative emotion representation. But, for negative identity representation, the representation of another person is chosen randomly, regardless of what pose or emotion it has. The goal of the triplet loss is to achieve,

$$d(f(X_i^a), f(X_i^n)) > d(f(X_i^a), f(X_i^p)) + m \qquad (6)$$

The optimal state for each single triplet loss is achieved when $d(f(X_i^a), f(X_i^p))$ is equal to zero, and $d(f(X_i^a), f(X_i^n))$ is greater than the predefined margin.

# 4. Results and Discussion

In this section, we initially present the benchmark dataset that we have employed to assess our proposed method. Then, we elaborate on the details of our implementation and introduce the metrics used in this paper. Finally, we demonstrate our experimental results and discussion.

## 4.1. Datasets

With the object of benchmarking our results, we have used the KDEF dataset. It is a publicly available dataset of 4900 face images, covering 140 unique identities. The images demonstrate face images with varying poses and emotion styles. Some samples of these datasets are shown in Figure 5.

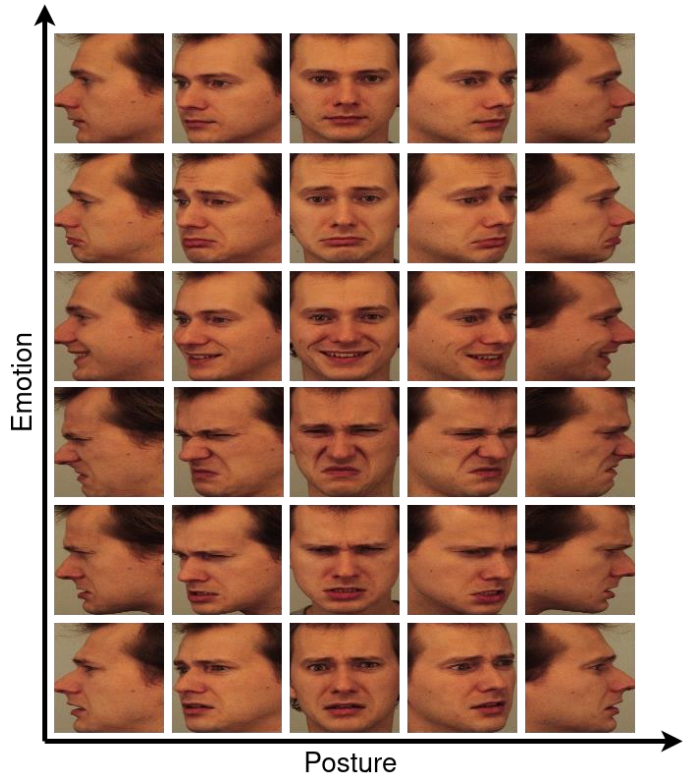

**Figure 5.** A few samples of the KDEF dataset. The KDEF dataset provides face images of 140 people in various postures and emotions.

## 4.2. Implementation details

We carried out our experiments on a machine with a Core i7-1165G7 @ 2.80GHz CPU with 64 Gigabytes of RAM and a GeForce RTX 2060 12 GB GPU. All models were implemented and trained using the Pytorch framework. Table 2 shows the hyperparameter setting.

**Table 2.** Details of the training procedure and the utilized FRLs. The hyperparameter settings are shown.

| FRL arch. | # epochs | Init. learning rate | Dropout rate | Triplet margin | ViT | | | | |
|---|---|---|---|---|---|---|---|---|---|
| | | | | | Embedding dim | FC dim | # Heads | # Layers | Patch size |
| MagFace | 255 | 0.001 | 0.4 | 10.0 | 256 | 256 | 4 | 4 | 8 |
| ArcFace | 320 | 0.001 | 0.4 | 10.0 | | | | | |
| CosFace | 157 | 0.001 | 0.05 | 10.0 | | | | | |

Furthermore, with the object of fairly evaluating the proposed FRA, we divided KDEF datasets based on identities with the following distributions:

- We randomly selected all samples from 99 identities, which nearly comprise 70.7 % of all identities in KDEF as our training data.
- Our validation data consists of samples from 11 identities, which represent approximately 7.8% of all identities in KDEF, and were selected randomly.
- Our testing data consisted of samples from 30 identities, which represent almost 21.5% of all identities in KDEF, and were selected randomly.

## 4.3. Experimental Results

This section details our comprehensive experimental results. Table 3 shows the achieved accuracy of the Support Vector Machine (SVM) (Cortes & Vapnik, 1995) classifier on the embeddings generated in three experiments. These experiments are: (1) Pre-augmentation accuracy with no augmentation, in which the training performs on the embeddings extracted using MagFace, ArcFace, and CosFace, and the testing accuracy is achieved on the testing partition of these embeddings (Train/Test split ratio is set 80/20), (2) Generated embeddings' accuracy, in which, we train SVM on augmented embeddings to demonstrate how much the proposed model can sustain the identity, posture, and emotion-related features without any degradation, (3) Post-augmentation accuracy, in which, we train SVM on augmented, while the test split is the same as a pre-augmentation experiment.

**Table 3.** Evaluation results of FRA. Pre-augmentation and post-augmentation accuracies show the effectiveness of FRA. Generated embeddings' accuracy denotes the sustainability of FRA.

| FRL | Target | (1) Pre-augmentation Accuracy (%) | (2) Generated embeddings' Accuracy (%) | (3) Post-augmentation Accuracy (%) |
|---|---|---|---|---|
| MagFace | Posture | 82.38 | 98.12 | 96.66 |
| | Identity | 86.19 | 93.61 | 95.71 |
| | Emotion | 44.76 | 92.43 | 99.04 |
| ArcFace | Posture | 89.12 | 99.3 | 97.9 |
| | Identity | 86.61 | 91.6 | 96.65 |
| | Emotion | 53.55 | 92.98 | 100 |
| CosFace | Posture | 99.12 | 99.91 | 99.12 |
| | Identity | 79.91 | 88.11 | 96.50 |
| | Emotion | 54.14 | 87.61 | 97.37 |

The utilization of an SVM classifier for classifying representations is justified for several reasons, as discussed in (Burbidge & Buxton, 2001). Firstly, SVM excels in scenarios of linear separability, where classes can be distinguished by a straight line or hyperplane in the feature space. MagFace representations, being highly discriminative, often occupy a feature space exhibiting favorable linear separability. Secondly, SVM's goal of maximizing the margin between the decision boundary and class data points enhances generalization and performance on new data. This aligns with the discriminative nature of MagFace representations, allowing SVM to establish distinct decision boundaries. Thirdly, SVM's efficacy with limited training samples is valuable, given that training deep learning models like MagFace demands substantial data. This efficiency aids computational and resource requirements. Lastly, the capacity of SVM to utilize the kernel trick allows for the classification of nonlinear data by implicitly mapping it to higher dimensions. This becomes beneficial when confronted with MagFace representations that are not linearly separable. The default Radial Basis Function (RBF) kernel is employed for the SVM classifier.

Based on Table 3, it is observed that our proposed approach improves the classification accuracy, not only identity-wise but also in terms of emotion and posture. For instance, SVM outputs 86.19% accuracy on the MagFace embeddings, but after the augmentation, this score goes up to 95.71% in Post-augmentation. The embeddings generated for the same data are significantly more representative. This improvement in representation leads to a 93.61% increase in classification accuracy when identifying the SVM outputs. This enhancement can also be validated for ArcFace and CosFace since our approach increases the accuracy in all three experiments. Also, FRA can improve the accuracy of SVM embeddings remarkably. In addition to improving the classification accuracy concerning the identities of the embeddings, FRA improves that pose and emotion-wise. Based on Table 5, the accuracy of the SVM classifier is increased from 86.19% for MagFace embeddings to 95.71% after augmentation. This improvement for ArcFace and Cosface is from 86.61% to 96.65% and from 79.91% to 97.37%, respectively.

Furthermore, our generated embeddings should ensure that they are linearly separable. The embeddings can be classified using a linear classifier such as SVM with a linear kernel. In our experiments, we used an SVM classifier with a linear kernel. Based on Table 5, we can deduce that FRA can improve the accuracy in Phase 2 and Phase 3 by a large margin, effectively enhancing the performance of SVM with a linear kernel.

As an extra experiment, we demonstrate that FRA is not reliant on any FR algorithms. We accomplish this by utilizing three different approaches: MagFace, ArcFace, and CosFace. After enhancing the embeddings produced by each of these algorithms, according to Table 5, the SVM classifier's classification accuracy experiences a noticeable improvement. This is proof that FRA does not depend on any specific FR algorithm, as it has all its requirements fulfilled.

Additionally, the reconstructed binary images produced by the autoencoder are showcased to validate the exceptional performance of our approach. These images are illustrated in Figure 6, showing the original image and the AE's output when dealing with MagFace embeddings, ArcFace embeddings, and CosFace embeddings.

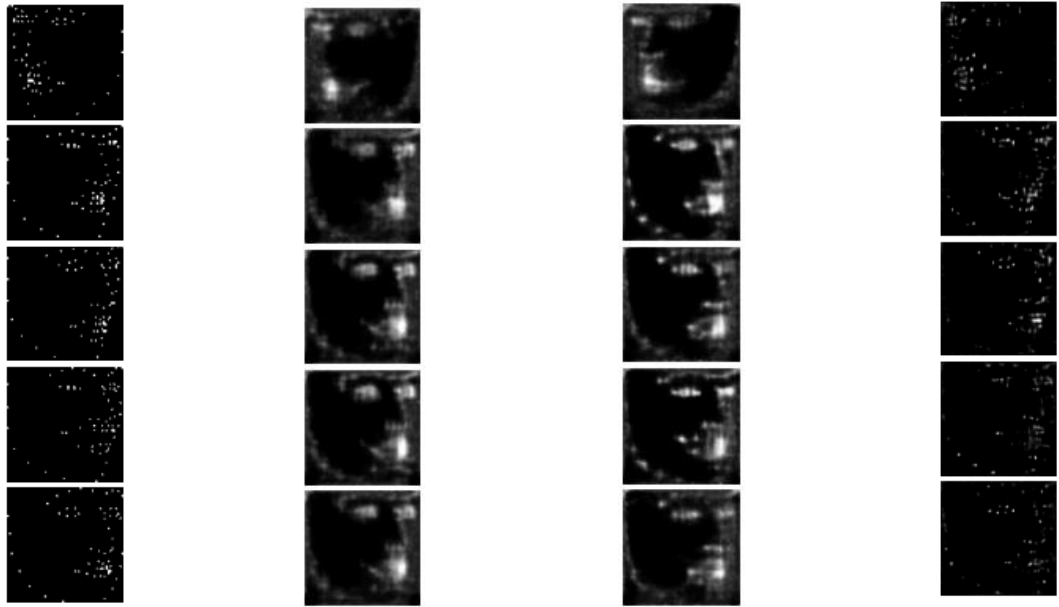

| Original landmarks | AE-generated landmarks for Magface FRL | AE-generated landmarks for Arcface FRL | AE-generated landmarks for Cosface FRL |

**Figure 6.** Some instances of reconstructed binarized facial landmark images. These reconstructed images indicate the performance of the AE for each FRL model. The CosFace model reconstructs most landmarks more precisely.

The training and validation loss in the training procedure of our proposed approach is shown in Figure 7. The training process of the proposed approach clearly converges smoothly. The smooth convergence occurs due to various reasons, namely normalizing the generated representations, using representation vectors with reasonably adjusted dimensions, and proper loss function.

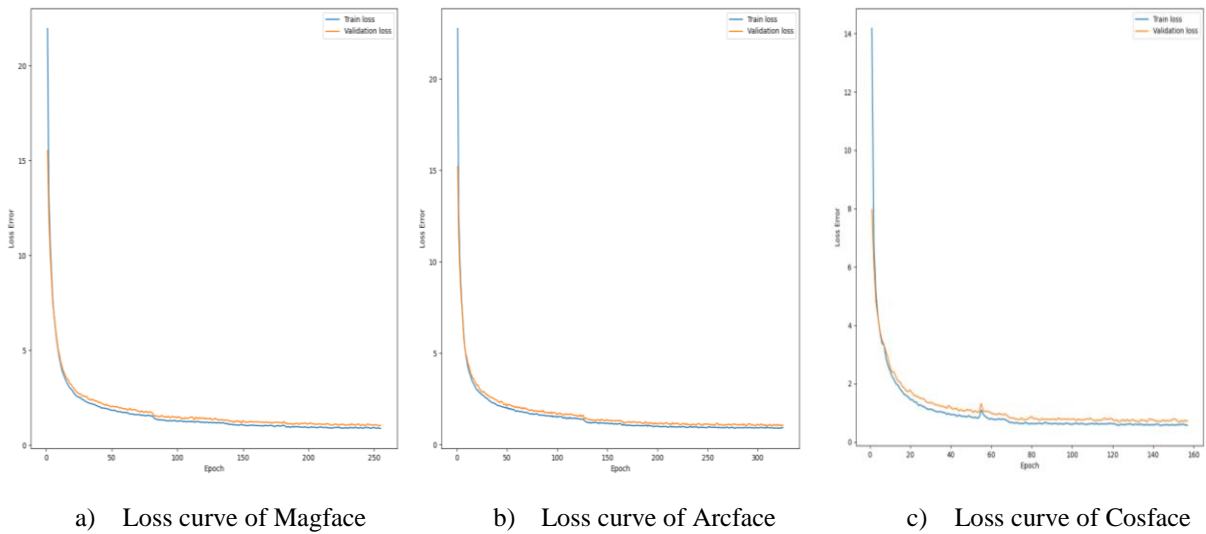

a) Loss curve of Magface    b) Loss curve of Arcface    c) Loss curve of Cosface

**Figure 7.** Total loss curves were achieved by various FRL methods.

In the context of our research investigation, we conducted a comprehensive analysis to assess the impact of data augmentation on the performance of the proposed approach. To quantitatively measure the efficacy of FRA, we meticulously compared the Receiver Operating Characteristic (ROC) curves before and after augmentation of all experiments for MagFace, ArcFace, and CosFace, as shown in Figure 8. Strikingly, our evaluation revealed a notable enhancement in the post-augmentation ROC curve when contrasted with the pre-augmentation counterpart. This significant enhancement highlights the efficacy of the augmentation process in enhancing the model's ability to discriminate. The visibly higher Area Under the Curve (AUC) values in the post-augmentation ROC curve directly highlight the augmentation's contribution to the model's heightened discriminatory power.

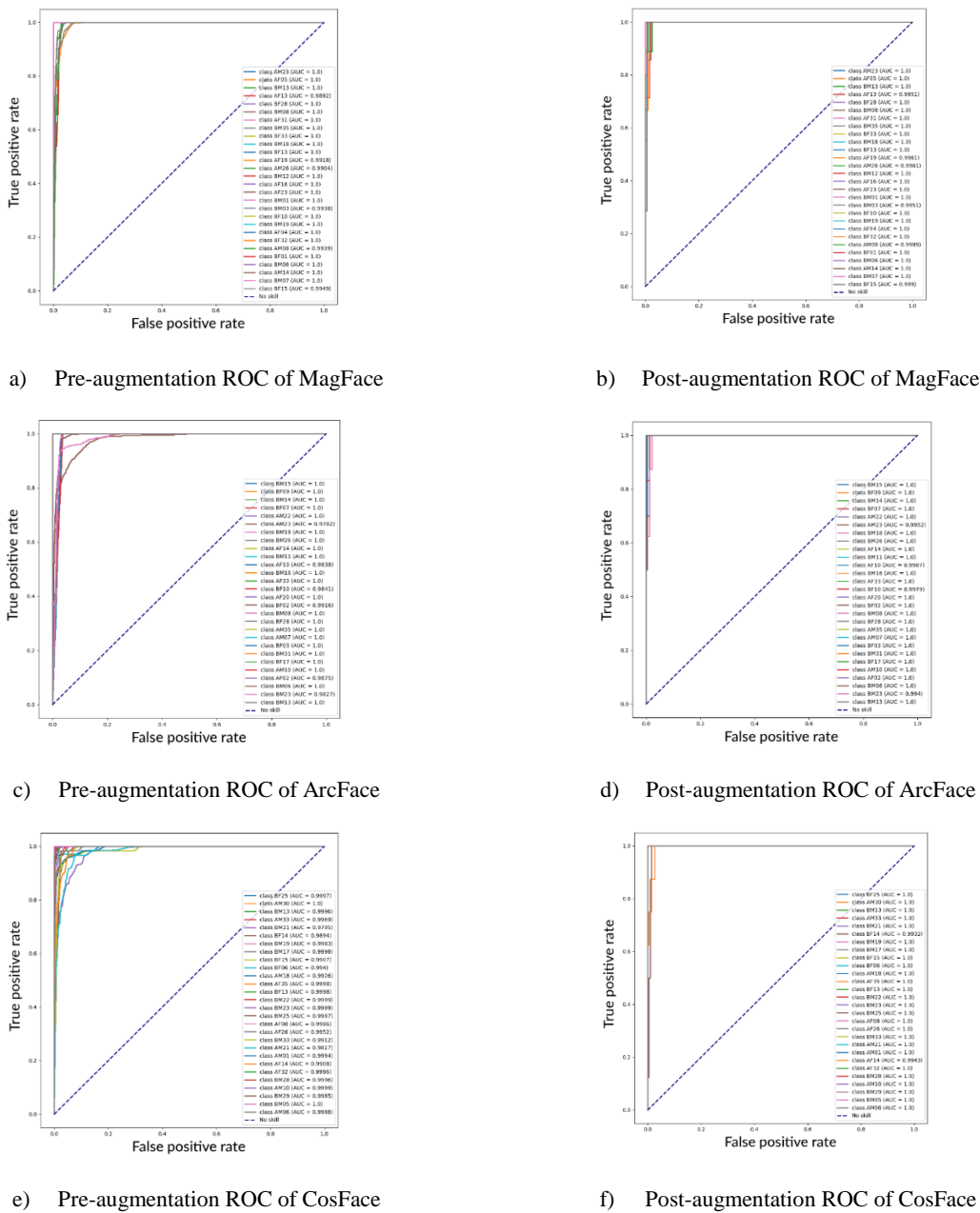

a)   Pre-augmentation ROC of MagFace  b)   Post-augmentation ROC of MagFace

c)   Pre-augmentation ROC of ArcFace  d)   Post-augmentation ROC of ArcFace

e)   Pre-augmentation ROC of CosFace  f)   Post-augmentation ROC of CosFace

**Figure 8.** ROC curves for MagFace, ArcFace, and CosFace. As the curves indicate, post-augmentation ROC curves (b, d, and f) have significant improvements in comparison to pre-augmentation ROC curves (a, c, and e), respectively.

Consequently, this outcome substantiates the practicality and utility of our augmentation strategy, offering valuable insights for advancing classification methodologies within the domain of machine learning research.

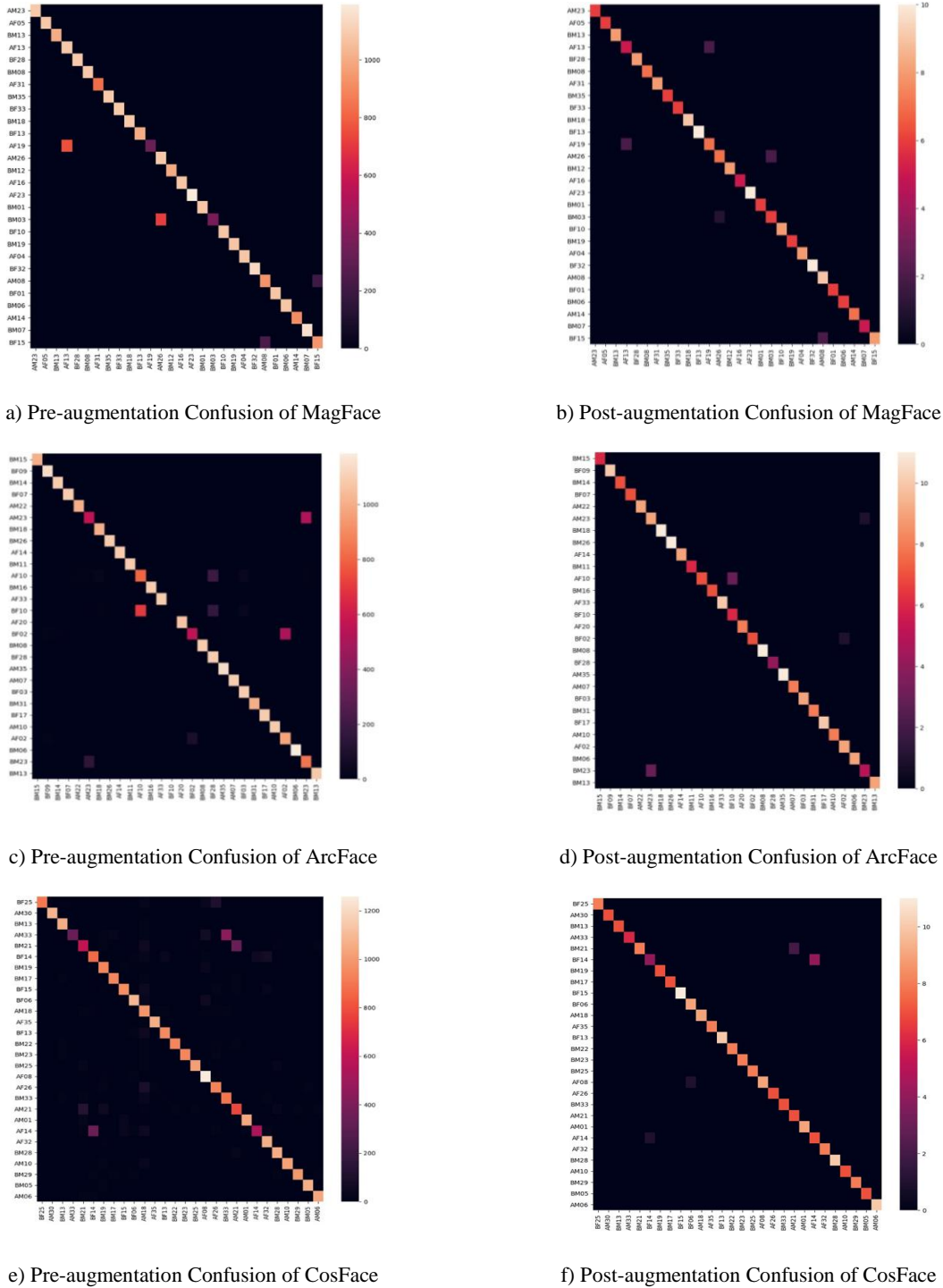

a) Pre-augmentation Confusion of MagFace

b) Post-augmentation Confusion of MagFace

c) Pre-augmentation Confusion of ArcFace

d) Post-augmentation Confusion of ArcFace

e) Pre-augmentation Confusion of CosFace

f) Post-augmentation Confusion of CosFace

**Figure 9.** Confusion matrices for MagFace, ArcFace, and CosFace. The confusion matrices show that the post-augmentation matrices (b, d, and f) have made significant improvements compared to the pre-augmentation matrices (a, c, and e), respectively.

In addition, to support the improvements brought about by the proposed approach, we compared the pre-augmentation and post-augmentation confusion matrices of all experiments for MagFace, ArcFace, and CosFace in Figure 9. Remarkably, our analysis revealed a discernible enhancement in the post-augmentation confusion matrix compared to its pre-augmentation counterpart. This improvement underscores the efficacy of the augmentation process in refining the model's classification accuracy. The reduction in misclassifications and the alignment of predicted labels with actual observations in the post-augmentation confusion matrix signify the efficacy of our approach. These findings provide empirical evidence of the benefits derived from data augmentation, affirming its utility in bolstering the model's discriminative capacity and its ability to generalize effectively. This outcome substantiates the relevance and practicality of our augmentation strategy, thereby contributing to the advancement of classification techniques in machine learning research.

## 4.4. Discussion

FR has captivated the minds of erudite specialists and intellectuals in the realm of biometric identification, bestowing upon them the exquisite advantages of a touchless, affable, and effortlessly adaptable nature. Although some state-of-the-art approaches presented in the literature have demonstrated remarkable performance, there is still a need to enhance such algorithms in real-world situations. For better handling such uncontrolled contexts, especially when we face a lack of data, DA techniques are introduced to increase the number of training samples by applying different manipulations. Classical techniques for altering images, such as rotation, skewing, flipping, blurring, and so on, as well as GAN-based ones that employ deep generative models and disentangled features to produce more authentically transformed facial images, have been extensively researched for the purpose of DA in the field of face recognition. However, these techniques have their drawbacks. Classical techniques mostly manipulate face images in a way that distorts their alignment, causing FR algorithms' performance when generating distinct representative embeddings dramatically decrease. In order to demonstrate this, we have conducted experiments using four distinct transformations: a horizontal flip, skewing, blurring, and notifying. We augmented the samples in the KDEF dataset and expanded the size of the training dataset to be four times larger than the original dataset. Additionally, we utilized various FR algorithms to generate embeddings. Then, we classified the embeddings using SVM concerning their identities. Table 4 details the results achieved by this experiment.

**Table 4.** Evaluation of MagFace, ArcFace, and CosFace using traditional augmentation techniques on the KDEF dataset. Post-augmentation accuracy scores achieved denote the ineffectiveness of these techniques for FR tasks.

| FR Algorithm | Accuracy Score | |
|---|---|---|
| | Pre-augmentation | Post-augmentation |
| MagFace | 20.49% | 18.54% |
| ArcFace | 20.12% | 17.01% |
| CosFace | 18.19% | 11.33% |

Table 4 shows that augmenting the face images using the classical approaches does not result in any improvement, and they degrade the quality of embeddings. For instance, for the MagFace algorithm, the accuracy obtained by SVM is decreased by 2% after applying DA.

Moreover, we conducted another experiment using three state-of-the-art generative-based algorithms, namely, CPM (Nguyen, Tran, & Hoai, 2021), AttGAN (He, Zuo, Kan, Shan, & Chen, 2019), and PSGAN

(W. Jiang et al., 2020). Following the previous experiments, we augmented the face images and obtained the classification accuracy before and after augmentation. Table 5 shows the results achieved by this experiment.

**Table 5.** Evaluation of augmentation techniques using GANs on the KDEF dataset. In the best case, the post-augmentation accuracy has increased a little, and in some cases, has caused a degradation in accuracy.

| Algorithm | Accuracy Score | |
| --- | --- | --- |
| | Pre-augmentation | Post-augmentation |
| CPM (W. Jiang et al., 2020) | 20.49 % | 17.82 % |
| AttGAN | 20.49 % | 26.70 % |
| PSGAN (Kemelmacher-Shlizerman, 2016) | 20.49 % | 23.40 % |

Based on Table 5, it can be claimed that these generative models do not contribute to classification accuracy improvement. Therefore, to address this issue, in this paper, we propose a new algorithm named FRA, which effectively augments the training data for FR algorithms. FRA operates using the original embeddings and modifies them in such a manner as to accurately reflect both the embedding's identity and the distinct postural information that is present in these representative embeddings. Results achieved by our extensive experiments indicate the efficacy of FRA in augmenting samples in the FR domain.

# 5. Conclusion

Since data scarcity is a common problem in deep learning-based solutions, it can be very challenging to build up FR systems that are robust enough to recognize face images with extreme diversity. In this paper, we proposed a method that augments the face data in latent space. The proposed method utilizes two major components, one of which is an autoencoder, and the other is a ViT-based model. The former encodes the binarized input images consisting of sparse facial landmarks into a latent space. The latter is used for extracting features from the combined embeddings from a pre-trained FRL approach and the autoencoder part of our model. Lastly, the output of the proposed model is an embedding representing the main identity with the same emotion but with a different posture. This way, we improved the classification accuracy by 9.52, 10.04, and 16.60, in comparison with the based models of MagFace, ArcFace, and CosFace, respectively. For future work, our objective is to study the effectiveness of the suggested approach in augmenting posture and emotion aspects within a latent space.

**Declaration of Competing Interest**
The authors declare that they have no known competing financial interests or personal relationships that could have appeared to influence the work reported in this paper.

**Acknowledgments**

Not applicable.

**Data Availability**

The data is from http://rasbt.github.io/mlxtend/, which is accessible.

**References**

Ali, W., Tian, W., Din, S. U., Iradukunda, D., & Khan, A. A. (2021). Classical and modern face recognition approaches: a complete review. *Multimedia Tools and Applications, 80*, 4825-4880.

Bai, Y., Lou, Y., Gao, F., Wang, S., Wu, Y., & Duan, L.-Y. (2018). Group-sensitive triplet embedding for vehicle reidentification. *IEEE Transactions on Multimedia, 20*, 2385-2399.

Bank, D., Koenigstein, N., & Giryes, R. (2020). Autoencoders. *arXiv preprint arXiv:2003.05991*.

Bhattacharyya, A., Chatterjee, S., Sen, S., Sinitca, A., Kaplun, D., & Sarkar, R. (2021). A deep learning model for classifying human facial expressions from infrared thermal images. *Scientific reports, 11*, 20696.

Bradski, G. (2000). The openCV library. *Dr. Dobb's Journal: Software Tools for the Professional Programmer, 25*, 120-123.

Burbidge, R., & Buxton, B. (2001). An introduction to support vector machines for data mining. *Keynote papers, young OR12*, 3-15.

Cao, Q., Shen, L., Xie, W., Parkhi, O. M., & Zisserman, A. (2018). Vggface2: A dataset for recognising faces across pose and age. In *2018 13th IEEE international conference on automatic face & gesture recognition (FG 2018)* (pp. 67-74): IEEE.

Carion, N., Massa, F., Synnaeve, G., Usunier, N., Kirillov, A., & Zagoruyko, S. (2020). End-to-end object detection with transformers. In *Computer Vision–ECCV 2020: 16th European Conference, Glasgow, UK, August 23–28, 2020, Proceedings, Part I 16* (pp. 213-229): Springer.

Chen, W., Chen, X., Zhang, J., & Huang, K. (2017). Beyond triplet loss: a deep quadruplet network for person re-identification. In *Proceedings of the IEEE conference on computer vision and pattern recognition* (pp. 403-412).

Chollet, F. (2021). *Deep learning with Python*: Simon and Schuster.

Cong, K., & Zhou, M. (2022). Face Dataset Augmentation with Generative Adversarial Network. In *Journal of Physics: Conference Series* (Vol. 2218, pp. 012035): IOP Publishing.

Cordonnier, J.-B., Loukas, A., & Jaggi, M. (2019). On the relationship between self-attention and convolutional layers. *arXiv preprint arXiv:1911.03584*.

Cortes, C., & Vapnik, V. (1995). Support vector machine. *Machine learning, 20*, 273-297.

Deng, J., Guo, J., Xue, N., & Zafeiriou, S. (2019a). Arcface: Additive angular margin loss for deep face recognition. In *Proceedings of the IEEE/CVF conference on computer vision and pattern recognition* (pp. 4690-4699).

Deng, J., & Zafeririou, S. (2019b). Arcface for disguised face recognition. In *Proceedings of the IEEE/CVF International Conference on Computer Vision Workshops* (pp. 0-0).

Duan, Y., Lu, J., & Zhou, J. (2019). Uniformface: Learning deep equidistributed representation for face recognition. In *Proceedings of the IEEE/CVF Conference on Computer Vision and Pattern Recognition* (pp. 3415-3424).

Em, Y., Gag, F., Lou, Y., Wang, S., Huang, T., & Duan, L.-Y. (2017). Incorporating intra-class variance to fine-grained visual recognition. In *2017 IEEE International Conference on Multimedia and Expo (ICME)* (pp. 1452-1457): IEEE.

Goodfellow, I., Bengio, Y., & Courville, A. (2016). *Deep learning*: MIT press.

Goodfellow, I., Pouget-Abadie, J., Mirza, M., Xu, B., Warde-Farley, D., Ozair, S., Courville, A., & Bengio, Y. (2020). Generative adversarial networks. *Communications of the ACM, 63*, 139-144.

Gupta, S., Thakur, K., & Kumar, M. (2021). 2D-human face recognition using SIFT and SURF descriptors of face's feature regions. *The Visual Computer, 37*, 447-456.

Hafiz, A. M., Parah, S. A., & Bhat, R. U. A. (2021). Attention mechanisms and deep learning for machine vision: A survey of the state of the art. *arXiv preprint arXiv:2106.07550*.

He, Z., Zuo, W., Kan, M., Shan, S., & Chen, X. (2019). Attgan: Facial attribute editing by only changing what you want. *IEEE transactions on image processing, 28*, 5464-5478.

Huang, G. B., Mattar, M., Berg, T., & Learned-Miller, E. (2008). Labeled faces in the wild: A database forstudying face recognition in unconstrained environments. In *Workshop on faces in'Real-Life'Images: detection, alignment, and recognition*.

Huang, Y., Wang, Y., Tai, Y., Liu, X., Shen, P., Li, S., Li, J., & Huang, F. (2020). Curricularface: adaptive curriculum learning loss for deep face recognition. In *proceedings of the IEEE/CVF conference on computer vision and pattern recognition* (pp. 5901-5910).

Jiang, D., Hu, Y., Yan, S., Zhang, L., Zhang, H., & Gao, W. (2005). Efficient 3D reconstruction for face recognition. *Pattern Recognition, 38*, 787-798.

Jiang, W., Liu, S., Gao, C., Cao, J., He, R., Feng, J., & Yan, S. (2020). Psgan: Pose and expression robust spatial-aware gan for customizable makeup transfer. In *Proceedings of the IEEE/CVF Conference on Computer Vision and Pattern Recognition* (pp. 5194-5202).

Kammoun, A., Slama, R., Tabia, H., Ouni, T., & Abid, M. (2022). Generative Adversarial Networks for face generation: A survey. *ACM Computing Surveys, 55*, 1-37.

Kemelmacher-Shlizerman, I. (2016). Transfiguring portraits. *ACM Transactions on Graphics (TOG), 35*, 1-8.

Kim, T., Cha, M., Kim, H., Lee, J. K., & Kim, J. (2017). Learning to discover cross-domain relations with generative adversarial networks. In *International conference on machine learning* (pp. 1857-1865): PMLR.

Kortli, Y., Jridi, M., Al Falou, A., & Atri, M. (2018). A comparative study of CFs, LBP, HOG, SIFT, SURF, and BRIEF for security and face recognition. In *Advanced Secure Optical Image Processing for Communications*: IOP Publishing.

Krizhevsky, A., Sutskever, I., & Hinton, G. E. (2017). Imagenet classification with deep convolutional neural networks. *Communications of the ACM, 60*, 84-90.

Lahasan, B., Lutfi, S. L., & San-Segundo, R. (2019). A survey on techniques to handle face recognition challenges: occlusion, single sample per subject and expression. *Artificial Intelligence Review, 52*, 949-979.

Li, P., Prieto, L., Mery, D., & Flynn, P. J. (2019). On low-resolution face recognition in the wild: Comparisons and new techniques. *IEEE Transactions on Information Forensics and Security, 14*, 2000-2012.

Li, Y., Guo, K., Lu, Y., & Liu, L. (2021). Cropping and attention based approach for masked face recognition. *Applied Intelligence, 51*, 3012-3025.

Lin, T., Wang, Y., Liu, X., & Qiu, X. (2022). A survey of transformers. *AI Open*.

Liu, W., Wen, Y., Yu, Z., Li, M., Raj, B., & Song, L. (2017). Sphereface: Deep hypersphere embedding for face recognition. In *Proceedings of the IEEE conference on computer vision and pattern recognition* (pp. 212-220).

Liu, Z., Luo, P., Wang, X., & Tang, X. (2015). Deep learning face attributes in the wild. In *Proceedings of the IEEE international conference on computer vision* (pp. 3730-3738).

Lu, Z., Jiang, X., & Kot, A. (2018). Deep coupled resnet for low-resolution face recognition. *IEEE Signal Processing Letters, 25*, 526-530.

Lundqvist, D., Flykt, A., & Öhman, A. (1998). Karolinska directed emotional faces. *Cognition and Emotion*.

Lv, J.-J., Shao, X.-H., Huang, J.-S., Zhou, X.-D., & Zhou, X. (2017). Data augmentation for face recognition. *Neurocomputing, 230*, 184-196.

Masi, I., ;'n-/*, A. T., Hassner, T., Sahin, G., & Medioni, G. (2019). Face-specific data augmentation for unconstrained face recognition. *International Journal of Computer Vision, 127*, 642-667.

Meng, Q., Xu, X., Wang, X., Qian, Y., Qin, Y., Wang, Z., Zhao, C., Zhou, F., & Lei, Z. (2021). PoseFace: Pose-invariant features and pose-adaptive loss for face recognition. *arXiv preprint arXiv:2107.11721*.
Meng, Q., Zhao, S., Huang, Z., & Zhou, F. (2021). Magface: A universal representation for face recognition and quality assessment. In *Proceedings of the IEEE/CVF Conference on Computer Vision and Pattern Recognition* (pp. 14225-14234).
Mohammadzade, H., & Hatzinakos, D. (2012). Projection into expression subspaces for face recognition from single sample per person. *IEEE Transactions on Affective Computing, 4*, 69-82.
Moschoglou, S., Papaioannou, A., Sagonas, C., Deng, J., Kotsia, I., & Zafeiriou, S. (2017). Agedb: the first manually collected, in-the-wild age database. In *proceedings of the IEEE conference on computer vision and pattern recognition workshops* (pp. 51-59).
Movshovitz-Attias, Y., Toshev, A., Leung, T. K., Ioffe, S., & Singh, S. (2017). No fuss distance metric learning using proxies. In *Proceedings of the IEEE international conference on computer vision* (pp. 360-368).
Nguyen, T., Tran, A. T., & Hoai, M. (2021). Lipstick ain't enough: beyond color matching for in-the-wild makeup transfer. In *Proceedings of the IEEE/CVF Conference on Computer Vision and Pattern Recognition* (pp. 13305-13314).
Niu, Z., Zhong, G., & Yu, H. (2021). A review on the attention mechanism of deep learning. *Neurocomputing, 452*, 48-62.
Park, N., & Kim, S. (2022a). Blurs behave like ensembles: Spatial smoothings to improve accuracy, uncertainty, and robustness. In *International Conference on Machine Learning* (pp. 17390-17419): PMLR.
Park, N., & Kim, S. (2022b). How do vision transformers work? *arXiv preprint arXiv:2202.06709*.
Peng, C., Wang, N., Li, J., & Gao, X. (2019). DLFace: Deep local descriptor for cross-modality face recognition. *Pattern Recognition, 90*, 161-171.
Rumelhart, D. E., Hinton, G. E., & Williams, R. J. (1985). Learning internal representations by error propagation. In: California Univ San Diego La Jolla Inst for Cognitive Science.
Sanakoyeu, A., Tschernezki, V., Buchler, U., & Ommer, B. (2019). Divide and conquer the embedding space for metric learning. In *Proceedings of the ieee/cvf conference on computer vision and pattern recognition* (pp. 471-480).
Schroff, F., Kalenichenko, D., & Philbin, J. (2015). Facenet: A unified embedding for face recognition and clustering. In *Proceedings of the IEEE conference on computer vision and pattern recognition* (pp. 815-823).
Sengupta, S., Chen, J.-C., Castillo, C., Patel, V. M., Chellappa, R., & Jacobs, D. W. (2016). Frontal to profile face verification in the wild. In *2016 IEEE winter conference on applicatio*ns of computer vision (WACV)* (pp. 1-9): IEEE.
Shen, Y., Luo, P., Yan, J., Wang, X., & Tang, X. (2018). Faceid-gan: Learning a symmetry three-player gan for identity-preserving face synthesis. In *Proceedings of the IEEE conference on computer vision and pattern recognition* (pp. 821-830).
Shi, Y., Yu, X., Sohn, K., Chandraker, M., & Jain, A. K. (2020). Towards universal representation learning for deep face recognition. In *Proceedings of the IEEE/CVF Conference on Computer Vision and Pattern Recognition* (pp. 6817-6826).
Smirnov, E., Oleinik, A., Lavrentev, A., Shulga, E., Galyuk, V., Garaev, N., Zakuanova, M., & Melnikov, A. (2019). Face representation learning using composite mini-batches. In *Proceedin*gs of the IEEE/CVF International Conference on Computer Vision Workshops* (pp. 0-0).
Sohn, K. (2016). Improved deep metric learning with multi-class n-pair loss objective. *Advances in neural information processing systems, 29*.
Song, L., Gong, D., Li, Z., Liu, C., & Liu, W. (2019). Occlusion robust face recognition based on mask learning with pairwise differential siamese network. In *Proceedings of the IEEE/CVF International Conference on Computer Vision* (pp. 773-782).

Srivastava, Y., Murali, V., & Dubey, S. R. (2020). A performance evaluation of loss functions for deep face recognition. In *Computer Vision, Pattern Recognition, Image Processing, and Graphics: 7th National Conference, NCVPRIPG 2019, Hubballi, India, December 22–24, 2019, Revised Selected Papers 7* (pp. 322-332): Springer.

Taigman, Y., Yang, M., Ranzato, M. A., & Wolf, L. (2014). Deepface: Closing the gap to human-level performance in face verification. In *Proceedings of the IEEE conference on computer vision and pattern recognition* (pp. 1701-1708).

Terhorst, P., Kolf, J. N., Damer, N., Kirchbuchner, F., & Kuijper, A. (2020). SER-FIQ: Unsupervised estimation of face image quality based on stochastic embedding robustness. In *Proceedings of the IEEE/CVF conference on computer vision and pattern recognition* (pp. 5651-5660).

Tian, Y., Peng, X., Zhao, L., Zhang, S., & Metaxas, D. N. (2018). CR-GAN: learning complete representat *arXiv:1806.11191*.

Touvron, H., Cord, M., Douze, M., Massa, F., Sablayrolles, A., & Jégou, H. (2021). Training data-efficient image transformers & distillation through attention. In *International conference on machine learning* (pp. 10347-10357): PMLR.

Tran, L., Yin, X., & Liu, X. (2017). Disentangled representation learning gan for pose-invariant face recognition. In *Proceedings of the IEEE conference on computer vision and pattern recognition* (pp. 1415-1424).

Ustinova, E., & Lempitsky, V. (2016). Learning deep embeddings with histogram loss. *Advances in neural information processing systems, 29*.

Vaswani, A., Shazeer, N., Parmar, N., Uszkoreit, J., Jones, L., Gomez, A. N., Kaiser, Ł., & Polosukhin, I. (2017). Attention is all you need. *Advances in neural information processing systems, 30*.

Vincent, P., Larochelle, H., Lajoie, I., Bengio, Y., Manzagol, P.-A., & Bottou, L. (2010). Stacked denoising autoencoders: Learning useful representations in a deep network with a local denoising criterion. *Journal of machine learning research, 11*.

Wang, H., Wang, Y., Zhou, Z., Ji, X., Gong, D., Zhou, J., Li, Z., & Liu, W. (2018). Cosface: Large margin cosine loss for deep face recognition. In *Proceedings of the IEEE conference on computer vision and pattern recognition* (pp. 5265-5274).

Wang, J., & Perez, L. (2017). The effectiveness of data augmentation in image classification using ////////////////////////////////////////////////////////////deep learning. *Convolutional Neural Networks Vis. Recognit, 11*, 1-8.

Wang, M., & Deng, W. (2021). Deep face recognition: A survey. *Neurocomputing, 429*, 215-244.

Wang, X., Wang, S., Wang, J., Shi, H., & Mei, T. (2019). Co-mining: Deep face recognition with noisy labels. In *Proceedings of the IEEE/CVF International Conference on Computer Vision* (pp. 9358-9367).

Wolf, L., Hassner, T., & Maoz, I. (2011). Face recognition in unconstrained videos with matched background similarity. In *CVPR 2011* (pp. 529-534): IEEE.

Wong, S. C., Gatt, A., Stamatescu, V., & McDonnell, M. D. (2016). Understanding data augmentation for classification: when to warp? In *2016 international conference on digital image computing: techniques and applications (DICTA)* (pp. 1-6): IEEE.

Wu, C.-Y., Manmatha, R., Smola, A. J., & Krahenbuhl, P. (2017). Sampling matters in deep ;l/embedding learning. In *Proceedings of th
e IEEE international conference on computer vision* (pp. 2840-2848).

Xiao, Q., Luo, H., & Zhang, C. (2017). Margin sample mining loss: A deep learning based method for person re-identification. *arXiv preprint arXiv:1710.00478*.

Yan, C., Meng, L., Li, L., Zhang, J., Wang, Z., Yin, J., Zhang, J., Sun, Y., & Zheng, B. (2022). Age-invariant face recognition by multi-feature fusionand decomposition with self-attention. *ACM Transactions on Multimedia Computing, Communications, and Applications (TOMM), 18*, 1-18.

Yang, F., Yang, H., Fu, J., Lu, H., & Guo, B. (2020). Learning texture transformer network for image super-resolution. In *Proceedings of the IEEE/CVF conference on computer vision and pattern recognition* (pp. 5791-5800).

Ye, L., Rochan, M., Liu, Z., & Wang, Y. (2019). Cross-modal self-attention network for referring image segmentation. In *Proceedings of the IEEE/CVF conference on computer vision and pattern recognition* (pp. 10502-10511).

Zhang, X., Fang, Z., Wen, Y., Li, Z., & Qiao, Y. (2017). Range loss for deep face recognition with long-tailed training data. In *Proceedings of the IEEE International Conference on Computer Vision* (pp. 5409-5418).

Zheng, T., & Deng, W. (2018). Cross-pose lfw: A database for studying cross-pose face recognition in unconstrained environments. *Beijing University of Posts and Telecommunications, Tech. Rep, 5*.

Zheng, T., Deng, W., & Hu, J. (2017). Cross-age lfw: A database for studying cross-age face recognition in unconstrained environments. *arXiv preprint arXiv:1708.08197*.

Zheng, Y., Pal, D. K., & Savvides, M. (2018). Ring loss: Convex feature normalization for face recognition. In *Proceedings of the IEEE conference on computer vision and pattern recognition* (pp. 5089-5097).